\documentclass{article}

\usepackage{microtype}
\usepackage{graphicx}
\usepackage{subfigure}
\usepackage{booktabs} 

\usepackage{hyperref}
\usepackage{amsmath}

\usepackage[accepted]{icml2020}

\icmltitlerunning{Open-domain goal-oriented dialogue agents}

\usepackage{verbatim} 

\begin{document}

\twocolumn[
\icmltitle{I love your chain mail! Making knights smile in a fantasy game world:
Open-domain goal-oriented dialogue agents}



\icmlsetsymbol{equal}{*}

\begin{icmlauthorlist}
\icmlauthor{Shrimai Prabhumoye$^*$}{cmu,fb}
\icmlauthor{Margaret Li$^*$}{fb}
\icmlauthor{Jack Urbanek}{fb}
\icmlauthor{Emily Dinan}{fb}
\icmlauthor{Douwe Kiela}{fb}
\icmlauthor{Jason Weston}{fb}
\icmlauthor{Arthur Szlam}{fb}
\end{icmlauthorlist}

\icmlaffiliation{cmu}{Language Technologies Institute, Carnegie Mellon University, Pittsburgh, PA, USA}
\icmlaffiliation{fb}{Faceboook AI Research, NY, USA}

\icmlcorrespondingauthor{Shrimai Prabhumoye}{sprabhum@cs.cmu.edu}

\icmlkeywords{Machine Learning, ICML}

\vskip 0.3in
]



\printAffiliationsAndNotice{\icmlEqualContribution} 

\begin{abstract}
Dialogue research tends to distinguish between chit-chat and goal-oriented tasks. While the former is arguably more naturalistic and has a wider use of language, the latter has clearer metrics and a straightforward learning signal. Humans effortlessly combine the two, for example engaging in chit-chat with the goal of exchanging information or eliciting a specific response. Here, we bridge the divide between these two domains in the setting of a rich multi-player text-based fantasy environment where agents and humans engage in both actions and dialogue. Specifically, we train a goal-oriented model with reinforcement learning against an imitation-learned ``chit-chat'' model with two approaches:
 the policy either learns to pick a topic or learns to pick an utterance given the top-$K$ utterances from the chit-chat model.
We show that both models outperform an inverse model baseline and can converse naturally with their dialogue partner in order to achieve goals.
\end{abstract}

\section{Introduction}
\label{submission}

In the literature on artificial dialogue agents, 
 a distinction is often made between ``goal-oriented'' dialogue, where an agent is tasked with filling slots or otherwise obtaining or disseminating specified information from the user to help complete a task,
and open-domain ``chit-chat'', where an agent should imitate human small talk. 
Modeling goal-oriented dialogue can have advantages over chit-chat imitation as it gives clearer metrics of success and perhaps more meaningful learning signals; but goal-oriented dialogue data is often more specialized, covering only a narrow slice of natural language.
Current goal-oriented datasets study settings like booking restaurants or airline tickets, or obtaining weather information, as standalone tasks~\citep{raux2005let, henderson2014second, bordes2016learning, asri2017frames, budzianowski2018multiwoz}.
Chit-chat agents, by contrast, might focus on coarse statistical regularities of dialogue data without accurately modeling the underlying ``meaning''; but the  data often covers a much wider space of natural language. 
For example, Twitter or Reddit chit-chat tasks \citep{li2016persona,yang2018learning,mazare2018training} cover a huge spectrum of language and diverse topics.
Chit-chat and goal-oriented dialogue are not mutually exclusive: when humans engage in chit-chat, their aim is to exchange information, or to elicit specific responses from their partners. 
Modeling such goals, however, is made difficult by the fact that it requires large amounts of world knowledge, and that goals in real life are implicit.

In this work, we introduce a family of tasks that bridge the divide between goal-oriented and chit-chat dialogue, combining clearer metrics and learning signals on the one hand, with the richness and complexity of situated but open-domain natural language on the other.  
The tasks are set in a multi-player text-based fantasy environment \citep{urbanek2019learning} with grounded actions and reference objects. 
Given a particular character to play in a particular scenario (location, set of objects and other characters to interact with), an agent should conduct open-ended dialogue with the aim of persuading their dialogue partner to execute a specified action. 
The action could be an emote (smile, laugh, ponder, etc), or a game action (wear chain mail, drink mead, put glass on table, etc). 
The richness of the environment means that there are a huge set of possible tasks and scenarios in which to achieve a wide range of actions. 
We plan to make our entire setup, code and models publicly available.

We train a variety of baseline models to complete the task.  We compare agents trained to imitate human actions given a goal (an ``inverse model'') 
to two different RL approaches: optimizing actions with latent discrete variables (topics), or via rewarding actions sampled from the model (via the top-$K$ outputs).
We show that both types of RL agent are able to learn effectively, outperforming
the inverse model approach or the chit-chat imitation baseline, and can converse naturally with their dialogue partner to achieve goals.


In short, our main contributions are: a new family of tasks that combines goal-oriented dialogue and chit-chat in a rich, fully realized  environment, and the results and analysis of scalable RL algorithms and behavioral-cloning models (and simple heuristic methods) on these tasks.

\begin{figure*}
    \centering
    \includegraphics[width=\textwidth]{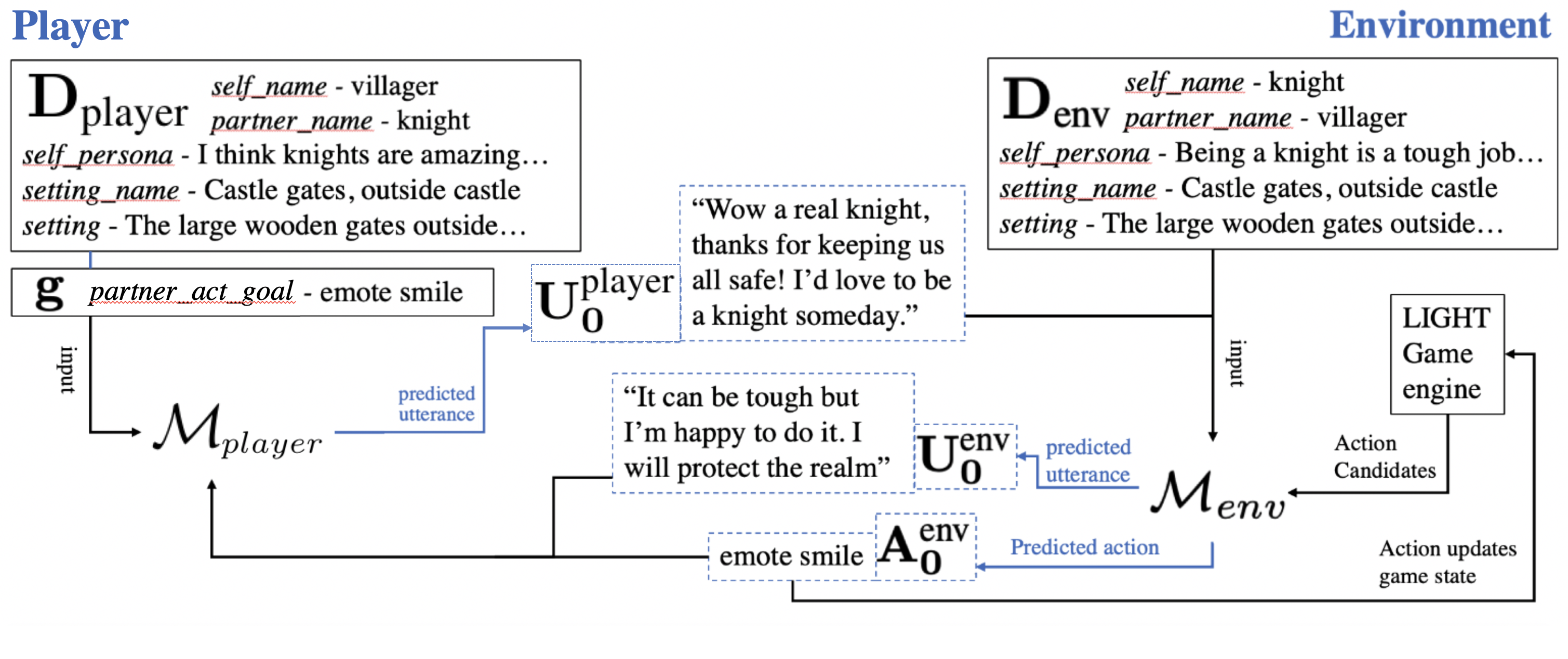}
    \vspace{-10mm}
    \caption{Example interaction in the described task setup (single turn). Here the RL agent $\mathbf{\mathcal{M}}_{player}$ would receive a reward as the environment agent $\mathbf{\mathcal{M}}_{env}$ took the desired action $\mathbf{g}$.} 
    \label{fig:rl-env-setup}
\end{figure*}

\section{LIGHT Game Environment} 
\label{sec:env}

We work in the LIGHT game environment \citep{urbanek2019learning}, which is a multi-user text-based game. 
Characters 
can speak to each other via free text, send emote actions like \emph{applaud}, \emph{nod} or \emph{pout} (22 emote types in total), and take actions to move to different locations and interact with objects (e.g. {\em get cutlery}, {\em put cutlery in drawer}, etc.), see Appendix \ref{sec:action_set} for a full list of game actions. 

The game engine itself is formally defined as a graph, where each location, object and character is a node, and they are connected by labeled edges, for example {\em contained-in}, {\em path-to} or {\em has-property}. 
Actions in the game result in changes in state of the graph. 
To a player (agent) a local view of the graph can be seen expressed as text, as are the game actions and changes of state. 
This text then naturally interleaves with the dialogue utterances of the speakers as well to form an input context sequence from which a character can base their subsequent actions.
See Appendix Figure \ref{example-light-episode} for an example episode of interactions between two humans in a given environment. 

To make the world and its textual descriptions, LIGHT consists of a large set of human-written game locations, characters, and objects, all based within a fantasy medieval setting.  
Their names, descriptions and properties were crowd-sourced, yielding a total of 663 locations, 1755 characters, and 3462 objects. 
They range from beaches with crabs and seaweed to crypts with archaeologists and coffins, yielding an extremely rich environment for agents to learn within.

Crowdworkers were then asked to play the role of characters 
within the game. 
This involved them making utterances, game actions and emotes, while interacting with each other (in pairs). 
The resulting gameplay data consists of  10,777 episodes with an average of 18.3 actions each of rich human play. 
These are split into train (8538), validation (500) and test (1739) portions, the latter being split into new episodes in existing settings (test seen, 1000) and completely new settings (test unseen, 739).
Players were not given specific goals, but instead asked to play the role convincingly of the character given, during play some of them effectively defined their own goals during the interactions,  see Appendix Figure \ref{example-light-episode}. 
Existing work \cite{urbanek2019learning} does not consider using this data to learn  goal-based tasks, but instead has only used this for chit-chat and action imitation learning. 



\section{Tasks}
The tasks we introduce in this work involve achieving open-domain goals during interaction between two agents in a given LIGHT scenario. 
One of the agents, which we will call the ``environment agent'' and write in symbols as $\mathbf{\mathcal{M}}_{env}$, together with the game engine, effectively functions as an environment for the other agent, which we will write $\mathbf{\mathcal{M}}_{player}$.
We assume that the environment agent is fixed; in this work it will be a model trained via behavioral cloning from human-human interaction data. 
$\mathbf{\mathcal{M}}_{player}$ must conduct open-ended dialogue such that a given goal action is executed in the future by the environment agent.



More formally: the two agents $\mathbf{\mathcal{M}}_{env}$ and $\mathbf{\mathcal{M}}_{player}$  are given their views of the scenario ($\mathbf{D}_{\text{env}}$ 
and  $\mathbf{D}_{\text{player}}$ respectively). 
These consist of the setting name, scenario description, character names, and their own persona, all described as a sequence of text (see Fig \ref{fig:rl-env-setup}).
Note that each agent can only access their own persona but not the persona of the partner with whom they are conversing, but they do know the name of their partner.   
Denote by $t$ the time-step of the environment, $\mathbf{U^{\text{player}}_t}$ and $\mathbf{U^{\text{env}}_t}$ the utterances of the agents $\mathbf{\mathcal{M}}_{player}$ and $\mathbf{\mathcal{M}}_{env}$ respectively, and denote by $\mathbf{A^{\text{env}}_t}$ the environment  actions by $\mathbf{\mathcal{M}}_{env}$.   
Hence the interaction sequence looks like 
\begin{multline} \mathbf{S}_t =  [\mathbf{U^{\text{player}}_0}, (\mathbf{U^{\text{env}}_0}, \mathbf{A^{\text{env}}_0}), \mathbf{U^{\text{player}}_1} ,
(\mathbf{U^{\text{env}}_1}, \mathbf{A^{\text{env}}_1})
, \\ \ldots, \mathbf{U^{\text{player}}_n}, (\mathbf{U^{\text{env}}_n}, \mathbf{A^{\text{env}}_n})]. 
\end{multline}
The agent $\mathbf{\mathcal{M}}_{player}$ is additionally given a persuasion goal $\mathbf{g}$ to achieve.
That is, the objective of $\mathbf{\mathcal{M}}_{player}$ is for $\mathbf{\mathcal{M}}_{env}$ to take the action $\mathbf{g}$.
An episode ends when $\mathbf{A}^{\text{env}}_t ==  \mathbf{g}$ or when $n$ becomes larger than a set number of turns.

\paragraph{Goals} We experiment separately with two different types of goals:
game actions and emote actions.
We use the  same train, valid, test (seen and unseen) split of the original human-human LIGHT episodes, assign roles $\mathbf{\mathcal{M}}_{player}$  and $\mathbf{\mathcal{M}}_{env}$ randomly, and randomly pick an action by $\mathbf{\mathcal{M}}_{env}$ that occurs in the episode  as the goal. We can then present the corresponding setting to our agents in order to form a new interaction, but within the same scenario and with a goal that was naturally desirable and achievable within that setting.

In our  experiments, $\mathbf{\mathcal{M}}_{player}$ only speaks, it does not perform game or emote actions. 
This was chosen in order to study grounded dialogue between agents;
 it guarantees that the player cannot  force the goal to be reached by performing actions itself. It has to produce appropriate utterances $\mathbf{U}^{\text{RL}}$ such that $\mathbf{\mathcal{M}}_{env}$ eventually takes the action $\mathbf{g}$.

\paragraph{Observations} The state observation $\mathbf{\mathcal{O}}_t = (\mathbf{D}_{\text{player}}, \mathbf{S}_{t-1}, \mathbf{g})$ at time $t$ given to a model consists of the agent's setting description ($\mathbf{D}_{\text{player}}$), the utterance and action history up to that time step ($\mathbf{S}_{t-1}$), and the agent's goal ($\mathbf{g}$).
Our models for $\mathbf{\mathcal{M}}_{player}$ consume $\mathbf{\mathcal{O}}_t$ as a flattened sequence of tokens, and return a dialogue utterance $\mathbf{U^{\text{player}}_t}$.
Each structured component is represented in the flattened sequenced separated by a special token denoting the types,  e.g. names, settings, etc., 
see Fig. \ref{fig:rl-env-setup}.

\subsection{Reinforcement learning formulation}
Our task set-up can be easily framed as a Markov decision process.  Because the entire history and goal is given to $\mathbf{\mathcal{M}}_{player}$, the environment is Markovian.    For the reward, we can give a terminal reward of +1 only if the goal $g$ is achieved and 0 otherwise, i.e, it is +1 if the environment agent takes the goal action  $g$.  The episode ends after
$n$  steps. In our experiments we consider $n=1$ and $n=3$.

When we formulate our tasks as a reinforcement learning problem, we will also refer to  $\mathbf{\mathcal{M}}_{player}$ as the ``RL agent''.

\section{Models}
\label{sec:models}
In this section we describe the models for $\mathbf{\mathcal{M}}_{env}$ and $\mathbf{\mathcal{M}}_{player}$.  In this work these are retrieval models, using the LIGHT dialogue training corpus as candidates (111k utterances). We leave generative models to future work. 

\paragraph{Base agent architecture} For all our models we adopt the same base architecture,
which is a 12-layer bidirectional transformer 
\citep{vaswani2017attention} pre-trained on a
large dialogue corpus (Reddit, 174M examples), and then fine-tuned on our task.
To score retrieval candidates, we use a {\it bi-encoder} as in \citep{humeau2019real,urbanek2019learning}.  That is, two transformers are used, one to encode the context, and another to encode a candidate response, and a dot product between the first output vector of each scores the match.
To produce a dialogue utterance, we take the utterance with the largest score from the training set candidates (111k in this case).
The same procedure is followed for actions and emotes. For actions, the candidates are the set of admissible actions at that game state, which are provided by the game engine, for example {\em get apple} is only available in the candidate set if it is a valid action (an apple is present in the room). For emotes, all 22 candidates are always available.
To train the model, a cross entropy loss is used.
Similar to \citet{mazare2018training}, during training we consider the other elements of the batch as negatives. 

\paragraph{Environment agent}  The environment agent is the base agent described above, and stays fixed over episodes where an RL agent is trained. This helps guarantee our RL models stick to using the semantics of natural language (English) rather than so-called language drift of learning a new emergent language on the same tokens \citep{lee2019countering}.
  
\paragraph{RL agents} We design two RL approaches for our tasks - learn to pick the right latent discrete variables (topics) that lead to goal-achieving utterances $\mathbf{U^{\text{player}}_i}$; or learn to pick the correct $\mathbf{U^{\text{player}}_i}$ from the top $K$ candidates. 
These are described in more detail in Sections \ref{sec:clusters} and \ref{sec:topk}.  
We also discuss a baseline ``inverse'' model trained via behavioral cloning on the human-human data.

\subsection{Inverse model}
\label{sec:inverse}
We consider an inverse model, trained to imitate human actions given a goal, as both a baseline for comparing to RL models, and for producing weights from which we can fine-tune.
The inverse model consists of a bi-encoder, as described above, which takes as input 
an observation $\mathbf{\mathcal{O}}_t$, 
and outputs an utterance. We train it by extracting from the human-human game logs
training set (which does not have goals) every instance where a game action occurs at time $t$ in $\mathbf{S_t}$, that is where 
\begin{multline}
\mathbf{S_t} = [ (\mathbf{U^{\text{player}}_1}, \mathbf{A^{\text{player}}_1}),  (\mathbf{U^{\text{env}}_1}, \mathbf{A^{\text{env}}_1}), \ldots, \\
(\mathbf{U^{\text{player}}_t}, \mathbf{A^{\text{player}}_t}), (\mathbf{U^{\text{env}}_t}, \mathbf{A^{\text{env}}_t})],
\end{multline}
and where $\mathbf{A^{\text{env}}_t}$ is not null (no action that turn); note, $\mathbf{A^{\text{player}}_i}$ for $0<i\leq t$ or $\mathbf{A^{\text{env}}_i}$ for $0 < i< t$ might be null.
We then construct a training example for the inverse model with observation
$ (\mathbf{D_{\text{player}}}, \mathbf{g}=\mathbf{A^{\text{env}}_t}, \mathbf{S}_{t-1})$. 
i.e. setting the goal $\mathbf{g}$ to be 
$\mathbf{A^{\text{env}}_t}$, and with the desired action to be taken by the agent as  
$\mathbf{U^{\text{player}}_t}$. 
Here we use the subscripts  ``player'' and ``env'' just to mark the relative positions in the sequence, as all actions and utterances come from the human logs.  Note also that unlike the RL agents we train, the human in the player agent ``position'' can take game actions.

We can thus train this model in a supervised manner using a cross entropy loss
as described before.
This model does not learn a policy interactively, and hence might not learn to plan or strategize optimally for goal completion.  The data distribution it is trained on is different than the data distribution seen by the RL agents.
However, 
it  serves as a strong baseline. Further, when training our RL
agents, we initialize their weights to the weights of this model, and then fine-tune
from that point.

\subsection{Latent Discrete Variable (Topic) Model}
\label{sec:clusters}

Optimizing all the parameters of a large transformer architecture by RL is both 
incredibly costly in data efficiency and computing time, 
and is also known to have the problem
of language drift \citep{lee2019countering} -- that is, there is no guarantee after training
with self-chat that the models will output recognizable natural language utterances.
A solution to both problems is to train most of the parameters of the model with human-human language data, and then to either disentangle or only optimize some of the parameters with model self-chat \citep{yarats2017hierarchical}.

Here, we propose a straight-forward model for that purpose.
We assume an RL agent that consists of two components.

The first component $F_{C}(\mathbf{\mathcal{O}}) = 
P_C(T_s(\mathbf{\mathcal{O}}))$ maps from an observation
to a discrete variable with $C$ possible values. It consists of a chain of two functions: a transformer $T_s$ that takes in the observation, and outputs a state representation $\tilde{s}$, and a policy chooser $c = P(\tilde{s}) \in (1,\dots,C)$ which takes in the state representation and outputs the value of the discrete latent variable.

The second component $T_{\mathit{u}}(\mathbf{\mathcal{O}}, c)$ is an additional transformer that takes as input the observation as well as the output of the first component, and outputs a dialogue utterance. The entire model is thus the chain  $u = T_{\mathit{u}}(\mathbf{\mathcal{O}}, P_C(T_s(\mathbf{\mathcal{O}})))$.
We make this explicit decomposition so that we can train only part of the model with RL; note that the ``action'' trained via RL is choosing $c$, not outputting the final utterance.

\paragraph{Initial topics}
We first 
pre-train the transformer $T_s$ using the inverse model described in 
Section \ref{sec:inverse}, which produces a vectorial representation of a given observation. We then run $K$-means over the vectorial representations of all observations from the training set to provide the mapping to one of $C$ values, which represent dialogue topics, which we use as our initial function $P_C(\tilde{s})$.
These two functions together give us our initialization of $F_{C}$.
Table \ref{tab:clusters} shows the cluster ID and the topic denoted by that cluster along with the most representative sentences (closest to the center) for that cluster for a model trained with  $50$ topics. As we can see, the clusters learnt can be coherent about a topic.
We use the set of topics as a set of actions $\mathbf{\mathcal{A}}$ for our RL setup.

\begin{table*}
\centering
\footnotesize
\setlength\tabcolsep{6.5pt} 
\begin{tabular}{lll}
\toprule
\bf{CID} & \bf{Topic} & \bf{Representative Sentences} \\
\midrule
19 & animal sounds & `Meow! Purr!', `Bah-Buk! Tasty!', `Woof! Very!', `Bock! Bock!'  \\
12 & find the cost & `I would love some fruit. What are your prices?', `They are beautiful. \\
& & How much do the cost?', `It flows easily, how much are you selling it for?' \\
28 & prayer, & `Then your poor life is a sign from God for you to join us in the churchand serve him!',  `If you say so priest. \\
& God &  From now I will pray every night for wealth and good food!', `Continue to love, worship, and serve Him.' \\
45 & ask favor & `Yes but do you mind doing me a favor?', `Since I have helped you,  could you do me a favor?',\\
& &  `If I offer to solve your problem, what will  you personally do for me in return?'\\
\bottomrule   
\end{tabular}
\caption{Clusters learnt from the dialogue utterances (Clusters = 50). `CID' denotes the cluster ID.}
\label{tab:clusters}
\end{table*}

\paragraph{From $c$~to~$\mathcal{A}$}
Given our initial choice of $F_{C}$,
we can also pre-train  
$T_{\mathit{u}}$. 
We simply take our initial human-human training data,
and for each observation append the topic computed by $F_c$ to it. This allows our model to be able to generate an action (utterance) conditional on both an input and a topic.
We can now train a policy by RL that optimizes the topic at any given point in the episode. 

\paragraph{Policy training}

We keep the pre-trained portions of the model $T_u$ and $T_s$ fixed and during fine-tuning only optimize $P_C$. The cluster chooser $P_C$ 
is redefined (from the initial $K$-means)  to be an MLP network consisting of 2 layers.
A discrete action is sampled from a categorical probability distribution over the possible topics, given by $ \mathbf{c_t} \sim \text{Categorical}(\mathbf{h_t^2})$, where
$\mathbf{h^2_t} = \text{tanh} (\mathbf{W_2} \text{tanh} (\mathbf{W_1} \mathbf{s_t} + b_1)  + b_2)$.

The state vector $\mathbf{s_t}$ also encodes the goal $\mathbf{g}$ and thus, the policy is conditioned on the goal $\mathbf{g}$ of the agent.
Hence, the policy  can learn strategies that will result in picking actions at each time step $\mathbf{t}$ that will help the agent to achieve its goal $\mathbf{g}$.
As our RL agent can only choose topics, it cannnot redefine easily the meaning of words to cause language drift.
We use the Advantage Actor-Critic implementation {\bf{A2C}} \cite{pytorchrl} to train the policy and the value function in both this and the subsequently described Top-$K$ model. 



\subsection{Top-$K$ model}
\label{sec:topk}
The Top-$K$ model, related to  \cite{dulac2015deep}, is another approach to keeping the number of trainable parameters small. As above it keeps close to the base retrieval model to avoid drift.  It first uses the inverse model to get a context embedding $\tilde{s}$ from the observation, and a list of $K$ candidate utterance embeddings $v_1, ... v_K$ corresponding to utterances $u_1, ... u_K$.   
These are the encodings by the inverse model of the $K$ utterances it considers most likely given the context and goal.  

We form scores $t_i = (A\tilde{s} + b)^Tv_i$, and obtain a probability distribution over these $K$ candidates for our policy: 
\begin{equation}\label{eq:biencoder} \pi(u_i| \text{context}) = \text{softmax}(t_0, ... , t_K)(i).\end{equation}
Here the trainable parameters of the RL agent are the map $A$ and biases $b$.

Alternatively, we can train a small (2-layer) Transformer model 
$T_{w}$
that takes as input the set $\{\tilde{s}, v_1, ... v_K\}$.  Instead of a softmax over dot products $t_i$ as in \eqref{eq:biencoder}, we use the attention weights in the  last layer of $T_w$ above $\tilde{s}$ against the candidates as the distribution over the candidates for sampling an utterance.  In this case, the weights of $T_w$ are the trainable parameters of the RL agent.    We call the former model a policy ``bi-encoder'' (Top-$K$-Bi in tables) and the latter simply Top-$K$.

\section{Related work}


\paragraph{Chit-chat dialogue}
There is an increasing body of work in the domain of chit-chat, where the primary
approaches being currently tried are end-to-end neural approaches. They are typically large pre-trained and then fine-tuned transformers, either
generative or retrieval. Retrieval models work best, or match generative models, on a number of tasks \citep{zhang2018personalizing,dinan2018wizard,li2019acute}. 
Our work shares a commonality with these approaches in that the original LIGHT dialogue data
we use has no specified goals, and humans chit-chat together (and act). Thus, the conversations cover a rich number of diverse topics. In \citet{urbanek2019learning} models were trained in a 
similar fashion to chit-chat task models, and we adopt similar architectures here, but instead adapt them
to learn to pursue goals.

\begin{table*}[h!]
\centering
\footnotesize
\setlength\tabcolsep{7pt} 
\begin{tabular}{llrrrrrr}
\toprule
 & & \multicolumn{3}{c}{\bf{Test Seen}} & \multicolumn{3}{c}{\bf{Test Unseen}} \\
 \cmidrule(lr){3-5} \cmidrule(lr){6-8} 
& & ($n=1$)   & \multicolumn{2}{c}{($n=3$)} & ($n=1$)   & \multicolumn{2}{c}{($n=3$)}\\
 \cmidrule(lr){3-3} \cmidrule(lr){4-5} \cmidrule(lr){6-6} \cmidrule(lr){7-8} 
 \bf{Model} & \bf{Goal Type} & Reward &  Reward & Turns & Reward &  Reward & Turns  \\
\midrule
Random Utterance & game act & 0.183 & 0.349 & 2.54 & 0.161 & 0.344 & 2.57 \\
Inverse model (no goal) & game act & 0.185 & 0.345 & 2.55 & 0.160 & 0.345 & 2.57  \\
Inverse model & game act & 0.223 & 0.414 & 2.42 & 0.193 & 0.410 & 2.48  \\
Top-$K$ RL & game act &  \bf{0.402} &  0.537  & 2.18 & \bf{0.331} & 0.449 &  2.35 \\
Top-$K$-BE RL & game act &  0.327 &  0.491  & 2.26 & 0.278 & 0.442 &  2.34 \\
Topic RL & game act & 0.359 & \bf{0.561} & 2.15&   0.313 & \bf{0.496} & 2.26 \\

Top-$K$ RL (1-step 3x) & game act &-~~~~ & 0.526 & 2.14 &-~~~~ & 0.475 & 2.26 \\
Topic RL (1-step 3x) & game act & -~~~~ & 0.493 & 2.22 & -~~~~ & 0.479 & 2.29  \\

\midrule
Random Utterance & emote & 0.086 & 0.200 & 2.79 & 0.061 & 0.185 & 2.81 \\
Inverse model (no goal) & emote & 0.072 & 0.219 & 2.77 & 0.075 & 0.212 & 2.78 \\
Inverse model & emote     & 0.089 & 0.262 & 2.72  & 0.088 & 0.266 & 2.74 \\
Top-$K$ RL & emote      & 0.166 & 0.400 & 2.55 & 0.131 & 0.349 & 2.59 \\
Top-$K$-BE RL & emote   & 0.219 & \bf{0.485} & 2.46 & 0.171 & \bf{0.436} & 2.53 \\
Topic RL & emote         & \bf{0.247} & 0.482 & 2.43 & \bf{0.208} & 0.427 & 2.49 \\
Top-$K$ RL (1-step 3x) & emote & -~~~~  & 0.336 &	2.58 & -~~~~	 & 0.293 & 	2.65
\\
Topic RL (1-step 3x) & emote & -~~~~ &  0.406 &	2.42 & -~~~~	& 0.348 & 2.50
\\

\bottomrule   
\end{tabular}
\caption{Results on the test seen and  unseen environments for our models.} 
\label{tab:main_results}
\end{table*}
\begin{figure*}[h!]
    \centering
    \includegraphics[width=0.45\textwidth]{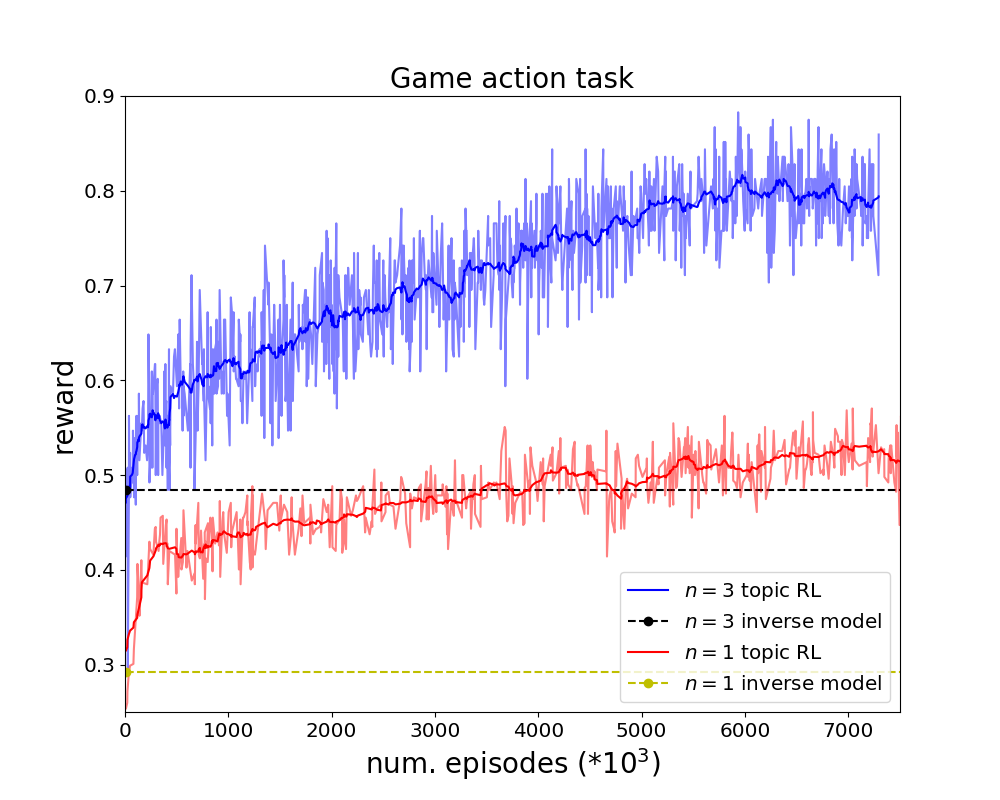}
    \includegraphics[width=0.45\textwidth]{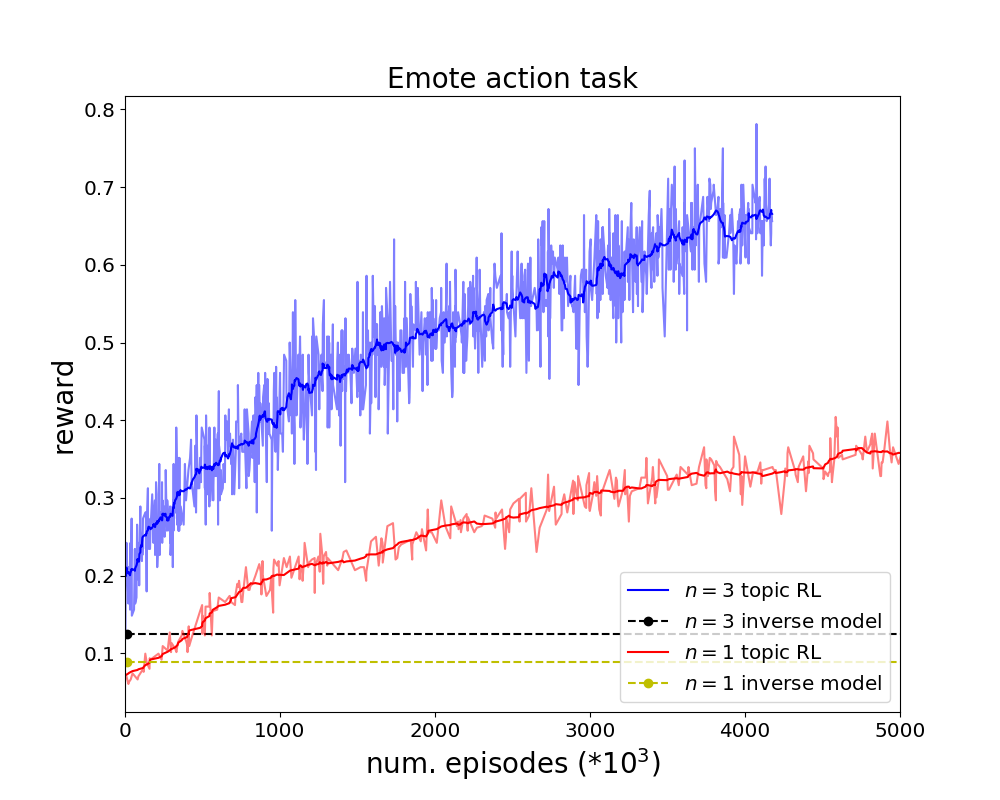}
    \caption{
    Topic RL model training  for $n=1$ and $n=3$ step goals for game actions (left) and emotes (right), comparing to the inverse model baselines.     Darker lines indicate smoothed plots. 
    Training using $8$ V100 machines took $\sim$2 weeks (1 step), $\sim$5 weeks (3 step).
    }
    \label{fig:rl-graph}
\end{figure*}

\begin{table*}[t!]
    \begin{footnotesize}
  \begin{center}
    \setlength\tabcolsep{4.5pt}
    \resizebox{0.9\textwidth}{!}{
     \begin{tabular}{llllllll}
        \toprule 
        \multicolumn{4}{l}{\textbf{Self:} guard ~~~~~ \textbf{Partner:} archer} & \multicolumn{4}{l}{\textbf{Self:} swimmer ~~~~~ \textbf{Partner:} turtles}  \\
        \cmidrule(lr){1-4} \cmidrule(lr){5-8} 
        \textbf{Persona:} & \multicolumn{3}{l}{I guard the castle. I guard the king.}   & \textbf{Persona:} & \multicolumn{3}{l}{ I am a huge fan of deep sea exploration,}\\
        & \multicolumn{3}{l}{I would kill to protect the royal family} & &  \multicolumn{3}{l}{but I take any chance I can get to go for a swim...} \\
        \cmidrule(lr){1-4} \cmidrule(lr){5-8} 
        \textbf{Setting:} & \multicolumn{3}{l}{The armory, Inside Tower.}   & \textbf{Setting:} & \multicolumn{3}{l}{ Bank, Swamp}\\
        & \multicolumn{3}{l}{The near top of the tower 6 feet before the very top.} & &  \multicolumn{3}{l}{This is a grassy area that surrounds much of the swamp.} \\
        & \multicolumn{3}{l}{Where the watchers keep their eye...} & &  \multicolumn{3}{l}{It's a plain field with some trees nearby along... } \\
        \cmidrule(lr){1-4} \cmidrule(lr){5-8} 
        $\mathbf{U^{\text{player}}_0}$ & \multicolumn{3}{l}{This is the armory!}   & $\mathbf{U^{\text{player}}_0}$ & \multicolumn{3}{l}{Just keep taking good care of your beautiful little}\\
        & \multicolumn{3}{l}{The king keeps the best weapons here.} & &  \multicolumn{3}{l}{turtle family! Your species is quite unique and I love} \\
        & \multicolumn{3}{l}{Take a look - } & &  \multicolumn{3}{l}{to see you about when I go for a swim.} \\
        \cmidrule(lr){1-4} \cmidrule(lr){5-8} 
        $\mathbf{U^{\text{env}}_0}$ & \multicolumn{3}{l}{Hello, I need to get into the palace to see the}   & $\mathbf{U^{\text{env}}_0}$ & \multicolumn{3}{l}{Well, thank you for that. Do you happen  to know}\\
         & \multicolumn{3}{l}{king. I think he might like to see these weapons.} & &  \multicolumn{3}{l}{ where my other turtle friend is? You     haven't captured } \\
          & \multicolumn{3}{l}{} & &  \multicolumn{3}{l}{any turtles have you?} \\
        \cmidrule(lr){1-4} \cmidrule(lr){5-8} 
        $\mathbf{A^{\text{env}}_0} $ & \multicolumn{3}{l}{get weapon}   & $\mathbf{A^{\text{env}}_0}$ & \multicolumn{3}{l}{hug swimmer}\\
        \bottomrule
        \\ 
        \toprule
         \multicolumn{4}{l}{{\textbf{Self:} townsperson ~~~~~ \textbf{Partner:} villager}} & \multicolumn{4}{l}{{\textbf{Self:} songbird ~~~~~ \textbf{Partner:} wasp}}  \\
        \cmidrule(lr){1-4} \cmidrule(lr){5-8} 
         \textbf{Persona:} & \multicolumn{3}{l}{We are the people who live in this  town.}   & \textbf{Persona:} & \multicolumn{3}{l}{ I fly high and bring beautiful music to the people.}\\
         & \multicolumn{3}{l}{We are common, and there are many...} & &  \multicolumn{3}{l}{I soar high and low going where the ... } \\
        \cmidrule(lr){1-4} \cmidrule(lr){5-8} 
        \textbf{Setting:} & \multicolumn{3}{l}{The Lagoon, Lake}   & \textbf{Setting:} & \multicolumn{3}{l}{ Meadow, Countryside}\\
        & \multicolumn{3}{l}{The Lagoon is a dark and mysterious place } & &  \multicolumn{3}{l}{Large clear outdoor meadow. Flowers of blue and } \\
        & \multicolumn{3}{l}{during the night hours. A lot of moss and lily...} & &  \multicolumn{3}{l}{white appearing in bunches here and there. The ...} \\
        \cmidrule(lr){1-4} \cmidrule(lr){5-8} 
        $\mathbf{U^{\text{player}}_0}$ & \multicolumn{3}{l}{It is cold up here. Would you like my coat}   & $\mathbf{U^{\text{player}}_0}$ & \multicolumn{3}{l}{Get out of here, wasp!}\\
        \cmidrule(lr){1-4} \cmidrule(lr){5-8} 
        $\mathbf{U^{\text{env}}_0}$ & \multicolumn{3}{l}{Oh yes please if I may. My shoe has become sodden}   & $\mathbf{U^{\text{env}}_0}$ & \multicolumn{3}{l}{You? Fly away from me? You're in my forest, bird.}\\
         & \multicolumn{3}{l}{from running to the market  I should love to dry it a bit.} & &  \multicolumn{3}{l}{I control this land.} \\
        \cmidrule(lr){1-4} \cmidrule(lr){5-8} 
        $\mathbf{A^{\text{env}}_0} $ & \multicolumn{3}{l}{remove Cloak}   & $\mathbf{A^{\text{env}}_0}$ & \multicolumn{3}{l}{hit a songbird}\\
        \bottomrule
      \end{tabular}
      }
      \caption{Example 1-step episodes where after the Topic RL agent's utterance $\mathbf{U^{\text{player}}_0}$ the environment agent's response action $\mathbf{A^{\text{env}}_0}$ was equal to the RL agent's goal $\mathbf{g}$. Our RL agent both makes natural utterances given the situation, and that elicit the desired goal.
      \label{example-rl-examples}}
  \end{center}
  \end{footnotesize}
\end{table*}

\begin{table}[h]
\centering
\footnotesize
\setlength\tabcolsep{7pt} 
\resizebox{0.48\textwidth}{!}{\begin{tabular}{llrrr}
\toprule
 & & \multicolumn{3}{c}{\bf{Train}}   \\
 \cmidrule(lr){3-5} 
& & ($n=1$)   & \multicolumn{2}{c}{($n=3$)} \\
 \cmidrule(lr){3-3} \cmidrule(lr){4-5}
 \bf{Model} & \bf{Goal} & Reward &  Reward & Turns  \\
\midrule
Top-$K$ RL & act &  0.677 &  0.752  & 1.72  \\
Topic RL &  act & 0.539 & 0.752 & 1.87 \\
Top-$K$ RL (1-st. 3x) &  act & -~~~~ & 0.737 & 1.62 \\
Topic RL (1-st. 3x) &  act & -~~~~ & 0.660 & 1.87  \\

\midrule
Top-$K$ RL & emote      & 0.498 & 0.668 & 2.13 \\
Topic RL & emote         & 0.483 & 0.612 & 2.22 \\
Top-$K$ RL \tiny(1st. 3x) & emote & -~~~~ & 0.587 &	1.96 	\\
Topic RL (1-st. 3x) & emote & -~~~~  & 0.570 &	1.99 	\\

\bottomrule   
\end{tabular}}
\caption{Results on the training environment for our models. 
\label{tab:train_results}
}
\end{table}

\begin{table*}[h] 
\centering 
\small
\setlength\tabcolsep{7pt} 
\begin{tabular}{lrrrrrrr} 
\toprule 
& & \multicolumn{2}{c}{\bf{1-Step}} & \multicolumn{2}{c}{\bf{1-Step 3x}} & 
\multicolumn{2}{c}{\bf{3-Step}}\\ 
\cmidrule(lr){3-4} \cmidrule(lr){5-6} \cmidrule(lr){7-8}
\bf{Verb} & Count & Topic &  Top-$K$ & Topic &  Top-$K$ & Topic &  Top-$K$ \\ 
\midrule 
get & 213 & 27.70 & 28.17 & 37.56 & 43.66 & \bf{44.13} & 40.85 \\ 
hit & 172 & 43.02 & 46.51 & 63.95 & 66.86 & 63.95 & \bf{75.58} \\ 
hug & 178 & 61.26 & 69.82 & 72.52 & 81.53 & 85.13 & \bf{85.56} \\ 
give & 136 & 33.09 & 41.91 & 50.00 & 54.41 & \bf{56.62} & 48.53 \\ 
remove & 127 & 9.45 & 13.39 & 22.83 & 22.83 & \bf{27.56} & 26.77 \\ 
steal & 55 & 47.27 & 50.91 & 63.64 & 63.64 & \bf{80.00} & 54.55 \\ 
drop & 27 & 0.00 & 0.00 & \bf{18.52} & \bf{18.52} & 7.41 & 7.41 \\ 
put & 25 & 0.00 & 0.00 & 8.00 & \bf{12.00} & 4.00 & 4.00 \\ 
eat & 10 & 30.00 & 10.00 & \bf{70.00} & 20.00 & 60.00 & 40.00 \\ 
wear & 10 & 0.00 & 0.00 & 20.00 & \bf{30.00} & 20.00 & 10.00 \\ 
drink & 3 & 33.33 & 33.33 & 33.33 & 33.33 & 33.33 & 33.33 \\ 
\bottomrule    
\end{tabular} 
\caption{Verb success in percentage on 1000 test seen episodes. The 3-step model performs best for high and medium frequency verbs.
\label{tab:verb_success} 
}
\end{table*}

\begin{table*}[h] 
\centering 
\small
\setlength\tabcolsep{7pt} 
\begin{tabular}{lrrrrrrrrr} 
\toprule 
& \multicolumn{3}{c}{\bf{1-Step}} & \multicolumn{3}{c}{\bf{1-Step 3x}} & 
\multicolumn{3}{c}{\bf{3-Step}}\\ 
\cmidrule(lr){2-4} \cmidrule(lr){5-7} \cmidrule(lr){8-10}
 & Topic &  Top-$K$ & Top-$K$-Bi & Topic &  Top-$K$ & Top-$K$-Bi & Topic &  Top-$K$ & Top-$K$-Bi  \\ 
\midrule 
1-step achievable & 0.452 & 0.505 & 0.407 & 0.616 & 0.647 & 0.587 & 0.686 & 0.664 & 0.620 \\ 
1-step unachievable & 0.000 & 0.005 & 0.005 & 0.044 & 0.058 & 0.044 & 0.068 & 0.049 & 0.078 \\ 
\bottomrule    
\end{tabular} 
\caption{Test seen breakdown by difficulty (1-step achievable or not). The 3-step models outperform the 1-step 3x models on both sets.
\label{tab:achievable_success} 
}  


\end{table*} 

\paragraph{Goal-oriented dialogue}
Traditional goal-oriented dialogue has focused on narrow tasks that would typically 
be useful for a dialogue-based assistant, for example
restaurant \citep{henderson2014second}, taxi, train, and hotel \citep{budzianowski2018multiwoz}
or trip \citep{asri2017frames} booking. Hence, each task typically focuses on a narrow slice of natural language and world knowledge for a specialized domain.
Earlier work focused on labeled state representations,
slot filling mechanisms and dialogue managers \citep{rieser2011reinforcement}, 
and more recent work has shifted to an end-to-end approach
\citep{bordes2016learning}, in line with chit-chat models, but still the two sets of tasks are rarely considered together, or by using the same methods.
Recently, \citet{tang2019target} used coarse-grained keywords as targets for open-domain chit-chat but in this work the target can be achieved when either the human or the agent uses the keyword in the response.

\paragraph{RL for dialogue}
The classical goal-oriented dialogue literature studies RL extensively \citep{singh2000reinforcement}.
Typically, they used RL 
to improve dialogue managers, which manage transitions
between dialogue states \citep{singh2002optimizing,pietquin2011sample,
rieser2011reinforcement,gasic2013pomdp,fatemi2016policy}. 
Recent works have focused more on end-to-end learning.
Some works have focused on self-play type mechanisms for end-to-end
reinforcement learning, where the reward is derived from the goal.
A related approach to ours is the negotiation task of \citet{lewis2017deal,yarats2017hierarchical}, which requires two agents to swap
3 item types (hats, balls, books) where the value of the items is different for 
the two agents, and derives their personal reward. In contrast, our setup encompasses a rich world of settings and characters -- with 3462 object types, and a corresponding large number of actions. This is reflected in the vocabulary size itself ($\sim$32,000 versus $\sim$2,000 in the negotiation tasks).
Other notable uses of RL in dialogue include within visual question answering
\citep{das2017learning},  in the domain of chit-chat where RL has been 
used to decrease repetitive and generic responses through the 
the use of self-play \citep{li2016deep}, and through human-bot conversation \citep{sankar2019deep}.

\paragraph{RL for language and games}
RL is used extensively for learning to play games, one of the most well known
examples being AlphaGo \citep{silver2016mastering}. Since then, language in games
has started to be more deeply explored, for example in graphical games such as Minecraft \citep{oh2017zero}, Real-time strategy war games \citep{hu2019hierarchical}, or in text adventure games \citep{narasimhan2015language,cote2018textworld}. The latter  are related to our setting.  However, those approaches use RL to optimize the set of actions given feedback in a {\em single-player} rather than multi-player game, so the text only refers to the environment, and there is no dialogue or actions from other agents.
Our work focuses on the latter.

\section{Experiments}

%
We compare our 
various models on the game action and emote action tasks.
We experiment with differing number of steps $n$ allowed to complete the goal, $n=1$ and $n=3$. 
Apart from the models described in Sec. \ref{sec:models}, we design two naive baselines to check the sanity of our environment models. 
The \textit{Random Utterance} model picks a random utterance from the set of all candidates and returns that response to the environment.
We also report results for the inverse model which does not get a goal to achieve.
Our main results for both seen and unseen test environments 
(\S\ref{sec:env})
are given in Table 
\ref{tab:main_results}. We report the average reward and for $n=3$ the average number of turns before completion. The results show clear improvements for our 
Topic RL (\S\ref{sec:clusters}) and Top-K RL (\S\ref{sec:topk})
compared to the inverse model and other baselines for each $n$,
for both game actions and emotes.

We show the training curves for Topic RL in Fig. \ref{fig:rl-graph}, reporting
rewards averaged over the batch (512 for $n=1$, and 128 for $n=3$). They show relatively
smooth improvements over time, with clear gains over the baseline. As a sanity check we
also tried, after training, to replace the Topic RL policy with 
random topic prediction, which yielded
poor results, e.g. 0.217 reward for $n=1$ test seen game actions. Our model is clearly learning appropriate
topic acts.

\paragraph{Example successful episodes}
We show examples of successful utterances, achieving goal actions in 
Fig. \ref{example-rl-examples} for a diverse range of scenarios, actions and language.
For example, for the guard's goal to encourage the archer to {\em get weapon} the 
Topic RL model utters ``This is the armory! The king keeps the best weapons here. Take a look'', which ends up leading to the desired action in the subsequent turn.
More examples (for $n=3$) are given in Appendix \ref{appsec:examples}.

\paragraph{Analysis of utterance choice}
To understand the semantics the models are learning that ground language to actions,
we visualize the top scoring utterances, averaged over their probabilities on the 1-step test set, broken down by verb type.
We observe a clear improvement in semantic connection for the Topic RL model over the inverse model. For example utterances such as ``Have a taste of this'' are highly scoring for {\em drink} goals, ``hmmnnnn.. this sure smells nice'' for {\em eat} goals, 
``Ew you vile beast, do not touch me!  I will have you removed'' for {\em hit} goals, and
``How I love being pampered by you, sweetheart'' for {\em hug} goals. Given there are $\sim$111,000 possible utterances in our setting, the model has clearly learned meaningful representations.  Detailed results are given in Appendix Tables \ref{tab:top_verb_utts_inv} and \ref{tab:top_verb_utts_topic}
for the inverse model and Topic RL model respectively.

\paragraph{Train vs. test performance}
We compare training performance of our models in Table \ref{tab:train_results}.
We see the same trends that models that performed better on test fit better on train  (e.g. Top-$K$  vs. Topic RL on 1-step tasks). 
Nevertheless, we do observe significant overfitting can occur, indicating that future  work could
explore either models that improve through better generalization, or by exploiting more
training data -- for example by self-play with more goals, rather than just using 
goals from human logs, as we have done here.

\paragraph{Model capacity}
We evaluate different values of $K$ or numbers of topics for Top-$K$ and Topic RL.
Full results are given in Appendix Table \ref{tab:k_value_exps}. They show that increasing
the capacity of both models improves performance up to 200 clusters or $K=200$, after which
performance saturates. However, $K=200$ (56.1\%) is substantially better than $K=50$ (47.7\%) on the 3-step task, for example.

\paragraph{Performance breakdown by goal}
We show the breakdown of test performance by goal type in Table \ref{tab:verb_success}
(splitting by verb type)
and Appendix Table \ref{tab:emote_success} (splitting by emote type). 
The results show that the easiest tasks are common actions with clear differentiation
such as {\em hug} (85\% success) and {\em hit} (75\%). Actions like {\em get}, {\em drop}, {\em give}
which are more confusable have somewhat lower numbers, with more rare actions (e.g. {wear})
faring worse. 

\paragraph{Performance breakdown by difficulty}

We can break down the test results into difficulty by considering in the 3-step task, which examples are 1-step achievable given the model's possible actions under the policy (i.e. the possible Top-$K$ utterances or Topic RL cluster choices), and reporting results separately. 
The results are given in Table \ref{tab:achievable_success}.
They show that non 1-step achievable goals are much harder, representing a significant challenge
to future systems.

\paragraph{1-step 3x baseline}
To investigate further the quality of our 3-step task models, 
we consider an additional baseline of taking a 1-step task trained model (Topic RL or Top-$K$) and applying it on  the 3-step task, which it has not been optimized for. 
The results in Table 
\ref{tab:main_results} show test results are inferior for this approach. Breaking
down further by goal type (Table \ref{tab:verb_success} and Appendix Table \ref{tab:emote_success}) shows that there are large improvements for the 3-step model on goals which are more often expressed in the data. Table \ref{tab:achievable_success} shows that 3-step models outperform the 1-step 3x models on both 1-step achievable and the harder 1-step unachievable goals.
Training performance (Table \ref{tab:train_results})  further
validates these results.

\paragraph{3-step task repeats}
We analyze the number of repeated utterances in an episode. The Topic RL model
repeats at least one utterance 25.8\% of the time, with  15.59\% utterances overall repeated.
The 1-step 3x baseline in comparison  repeats  37.3\% at least once, and 22.94\% on average. We note that repeating an utterance may possibly bring the desired goal in some cases, just as in real life.

\section{Conclusion}
In this paper, we investigate agents that can interact (speak or act) and
can achieve goals in a rich world with diverse language, bridging
the gap between chit-chat and goal-oriented dialogue.
We achieve this by defining a task for an agent, where the goal is for the other player to execute a particular action.
%
We explore two reinforcement learning approaches to solve this task, and compare them against a strong inverse model baseline.
We show that these approaches effectively learn dialogue strategies that lead to successful completion of goals, while producing natural chat.

Future work should develop improved agents that learn to act and 
speak in natural language at scale in our proposed open-domain task environment. This setup is exciting because it can be further generalized to richer and richer goal (game) states as we develop
models capable of them.


\nocite{langley00}

\bibliography{example_paper}
\bibliographystyle{icml2020}

\clearpage

\appendix


\onecolumn

\section{Additional Results}

\begin{table}[!htbp]
\centering
\setlength\tabcolsep{7pt} 
\begin{tabular}{lllrrrrrr}
\toprule
 & &  &  \multicolumn{3}{c}{\bf{Test Seen}} & \multicolumn{3}{c}{\bf{Test Unseen}} \\
 \cmidrule(lr){4-6} \cmidrule(lr){7-9} 
& & & ($n=1$)   & \multicolumn{2}{c}{($n=3$)} & ($n=1$)   & \multicolumn{2}{c}{($n=3$)}\\
 \cmidrule(lr){4-4} \cmidrule(lr){5-6} \cmidrule(lr){7-7} \cmidrule(lr){8-9} 
 \bf{Model} & \bf{Goal Type} & \# Clusters & Reward &  Reward & Turns & Reward &  Reward & Turns  \\
\midrule
Topic RL & game act & 50 &  0.324 & 0.477 & 2.31 & 0.277 &  0.470 & 2.24 \\
Topic RL & game act & 100 & 0.348 & 0.523 & 2.21 & 0.282 & 0.488 & 2.28 \\
Topic RL & game act & 200 & 0.359 & 0.561 & 2.15 & 0.313 & 0.496 & 2.26 \\
Topic RL & game act & 500 & 0.362 & 0.505  & 2.23 & 0.307 & 0.46 & 2.35 \\
Topic RL & game act & 1000 & 0.372 & 0.510 & 2.20 & 0.333 & 0.464 & 2.32  \\
\midrule
Top-$K$ RL & game act & 50 &  0.329 & 0.503 & 2.24 & 0.261 & 0.439 & 2.39 \\
Top-$K$ RL & game act & 100 & 0.370 & 0.521 & 2.12 & 0.292 & 0.468 & 2.33 \\
Top-$K$ RL & game act & 200 &  0.402 &  0.537  & 2.18 & 0.331 & 0.449 &  2.35 \\
Top-$K$ RL & game act & 500 & 0.402 & - & - & 0.299 & - & - \\
Top-$K$ RL & game act & 1000 & 0.426 & - & - & 0.337 & - & - \\


\bottomrule   
\end{tabular}
\caption{Results with different numbers of clusters (Topic RL) or candidates (Top-$K$ RL). Some experiments were not completed because of resource limitations.
\label{tab:k_value_exps}
}
\end{table}

\begin{table*}[!htbp] 
\centering 
\setlength\tabcolsep{7pt} 
\begin{tabular}{lrrrrrrr} 
\toprule 
& & \multicolumn{2}{c}{\bf{1-Step}} & \multicolumn{2}{c}{\bf{1-Step 3x}} & 
\multicolumn{2}{c}{\bf{3-Step}}\\ 
\cmidrule(lr){3-4} \cmidrule(lr){5-6} \cmidrule(lr){7-8}
\bf{Emote} & Count & Topic &  Top-$K$ & Topic &  Top-$K$ & Topic &  Top-$K$ \\ 
\midrule 
laugh & 109 & 20.18 & 11.01 & 32.11 & 20.18 & \bf{44.04} & 26.61 \\
smile & 106 & 31.13 & 13.21 & 58.49 & 37.74 & \bf{61.32} & 44.34 \\
ponder & 94 & 31.91 & 2.13 & 44.68 & 7.45 & \bf{59.57} & 24.47 \\ 
frown & 85 & 18.82 & 9.41 & 29.41 & 21.18 & \bf{34.12} & 24.71 \\ 
nod & 75 & 40.00 & 21.33 & 58.67 & 52.00 & \bf{84.00} & 56.00 \\
sigh & 67 & 55.22 & 4.48 & 82.09 & 14.93 & \bf{85.07} & 11.94 \\ 
grin & 63 & 4.76 & 1.59 & 25.40 & 12.70 & \bf{33.33} & 26.98 \\ 
gasp & 57 & 21.05 & 0.00 & \bf{33.33} & 0.00 & \bf{33.33} & 3.51 \\
shrug & 47 & 29.79 & 6.38 & 51.06 & 48.94 & \bf{59.57} & 48.94 \\
stare & 41 & 7.32 & 4.88 & \bf{26.83} & 17.07 & \bf{26.83} & 9.76 \\
scream & 40 & 17.50 & 20.00 & 25.00 & 25.00 & \bf{42.50} & 30.00 \\
cry & 32 & 12.50 & 28.13 & 18.75 & 50.00 & 43.75 & \bf{56.25} \\ 
growl & 27 & 40.74 & 37.04 & \bf{48.15} & 40.74 & 33.33 & 40.74 \\
blush & 26 & 3.85 & 19.23 & 11.54 & 50.00 & 19.23 & \bf{53.85} \\
dance & 24 & 37.50 & 29.17 & \bf{62.50} & 33.33 & \bf{62.50} & 33.33 \\
applaud & 23 & 17.39 & 0.00 & \bf{43.48} & 21.74 & 21.74 & 21.74 \\ 
wave & 19 & 21.05 & 21.05 & \bf{36.84} & 21.05 & 10.53 & 26.32 \\ 
groan & 17 & 5.88 & 0.00 & \bf{17.65} & 11.76 & 11.76 & 5.88 \\ 
nudge & 16 & 0.00 & 0.00 & 0.00 & 6.25 & 0.00 & \bf{12.50} \\
wink & 15 & 13.33 & 20.00 & 13.33 & 33.33 & 13.33 & \bf{53.33} \\ 
yawn & 11 & 0.00 & 0.00 & 0.00 & 18.18 & \bf{27.27} & \bf{27.27} \\ 
pout & 6 & 0.00 & 33.33 & 16.67 & \bf{66.67} & 16.67 & 16.67 \\ 
\bottomrule    
\end{tabular} 
\caption{Emote success in percentage on 1000 test seen episodes. The 3-step model performs best for high and medium frequency verbs.
\label{tab:emote_success} 
}
\end{table*}

\begin{table*}[!htbp]
\centering
\footnotesize
\setlength\tabcolsep{6.5pt} 
\begin{tabular}{p{0.5cm}p{0.5cm}p{13.8cm}}
\toprule
\bf{Verb} & \bf{count} & \bf{Top utterances} \\
\midrule
get & 213 & 'Why hello there, I haven;t seen you in awhile.', "Oh hello, I didn't expect to find anyone else here.", "Well hello there, wasn't expecting to see you here.", 'Wow! What a fine place this is.', "Oh, hello! I didn't see you all here.", 'Well hello there! I did not expect to see anyone here.', "Isn't this place so wonderful!?", 'I need some light.', 'So how is buisiness going?', '"Ah, what a long day we have ahead of us!"' \\
put & 25 & 'Why hello there, I haven;t seen you in awhile.', "Well hello there, wasn't expecting to see you here.", "Oh hello, I didn't expect to find anyone else here.", 'Wow! What a fine place this is.', 'Eerie. I must light a candle. And say a prayer', "Oh, hello! I didn't see you all here.", 'Well hello there! I did not expect to see anyone here.', "Isn't this place so wonderful!?", 'Greetings! How are my subjects doing this fine day?', 'Good morning. Someone needs to tend to this rickety rectory. I almost fell through the floor.'\\
drink & 3 & 'Eerie. I must light a candle. And say a prayer', 'It is a wonderful day to drink!  Time to get my drunk on!', 'I need another drink.', "Greetings m'lord! Cold day isn't it?", 'I am person just trying to enjoy the ambiance of this room', 'I need some light.', 'It appears you need some guidance.', 'Hello person! How are you on this fine evening?', 'Good evening good evening sir!  Can I help you?', "Well hello there, wasn't expecting to see you here." \\
eat & 10  & 'Why hello there, I haven;t seen you in awhile.', 'Hello bird, how are you doing?', 'Ahh, what a great day to nibble at the feet of humans.', 'I hope there is food in here.', 'Mmmm a human come into my territory. My lucky day indeed.', 'Ugh I am so tired of being used as food around here.', 'I am so delighted to not have to scavenge for food in the village.', 'WOW! So much food to eat here', '"Come here! I need to eat!"', 'man i hope i can find something to eat here' \\
steal & 55 & 'well what a fine mess i have gotten myself into this time', '*ARGH* you must let me out of this place.', 'I have seen you before! Thief what is it you think you will get today?', 'Wow, this lavoratory is filthy!', 'Hey, you there. Come here!', 'Hey, you over there! You look like you could use a little something I have.', 'Hello! You look as though you are in need of some of my wares.', 'It appears you need some guidance.', 'Why hello there, I haven;t seen you in awhile.', 'Enjoy!  You finally have a place of your very own.' \\
hit & 172 & 'Whatchit! You almost crushed me!', '*ARGH* you must let me out of this place.', 'Hey, you there. Come here!', 'well what a fine mess i have gotten myself into this time', 'Wow, this lavoratory is filthy!', 'You must bow before me.', 'Why are you in here! Back away from me or I will strike!', 'Why hello there, I haven;t seen you in awhile.', '"Come here! I need to eat!"', 'Ugh not another one of these beasts.'\\
hug & 222 & 'Why hello there, I haven;t seen you in awhile.', 'Minister! It is so good to see you!', "Well hello there, wasn't expecting to see you here.", "Oh hello, I didn't expect to find anyone else here.", "I'm so glad you're here with me", 'It is so nice and warm in here.', 'Wow! What a fine place this is.', 'I am so happy for this day.Even if is in this filthy place', "Oh, hello! I didn't see you are.", 'Hail, friend. How are things?'\\
wear & 10 & 'Why hello there, I haven;t seen you in awhile.', "Well hello there, wasn't expecting to see you here.", "Oh hello, I didn't expect to find anyone else here.", 'Wow! What a fine place this is.', 'Good afternoon sir! I did not expect to find you here.', 'Well hello there! I did not expect to see anyone here.', 'Why I did not expect to see you here, sir! Please join us.', 'Good evening good evening sir!  Can I help you?', 'It appears you need some guidance.', '"Ah, what a long day we have ahead of us!"'\\
drop & 27 & "Well hello there, wasn't expecting to see you here.", 'Why hello there, I haven;t seen you in awhile.', "Oh hello, I didn't expect to find anyone else here.", 'Wow! What a fine place this is.', "Oh, hello! I didn't see you all here.", 'Well hello there! I did not expect to see anyone here.', '"Ah, what a long day we have ahead of us!"', 'well what a fine mess i have gotten myself into this time', 'Oh, hello! I was just checking to see if anyone dropped these goblets. Ha, ha, ha.', 'So how is buisiness going?'\\
give & 136 & 'Why hello there, I haven;t seen you in awhile.', "Well hello there, wasn't expecting to see you here.", 'Wow! What a fine place this is.', "Oh hello, I didn't expect to find anyone else here.", 'Good evening good evening sir!  Can I help you?', "Isn't this place so wonderful!?", 'Well hello there! I did not expect to see anyone here.', "Oh, hello! I didn't see you all here.", 'Wow this is such a nice place.', 'I must get this place cleaned at once!' \\
remove & 127 & "Well hello there, wasn't expecting to see you here.", 'Why hello there, I haven;t seen you in awhile.', "Oh hello, I didn't expect to find anyone else here.", "Oh, hello! I didn't see you all here.", 'Wow! What a fine place this is.', 'Well hello there! I did not expect to see anyone here.', 'It appears you need some guidance.', 'Good evening good evening sir!  Can I help you?', 'Another hectic day in this place.', '"Ah, what a long day we have ahead of us!"' \\
\bottomrule   
\end{tabular}
\caption{Top utterances for each verb for the inverse model.}
\label{tab:top_verb_utts_inv}
\end{table*}

\begin{table*}[!htbp]
\centering
\footnotesize
\setlength\tabcolsep{6.5pt} 
\begin{tabular}{p{0.5cm}p{0.5cm}p{13.8cm}}
\toprule
\bf{Verb} & \bf{count} & \bf{Top utterances} \\
\midrule
get   &  213    & 'Here sir, I found this.', 'Oh hello there brothers! Why whose towel is this thats left all by its self?', 'How did this get here?', 'Meh. Whats this you have here?', "What is this? Is this someone's head?!", "Thank you, sir. What's with all this silk?", 'What is this here?', 'It looks like there is something missing!', "Oh, look, somethin' shinny", 'what is this ston slab' \\
put  &   25   &   'How did this get here?', 'Oh hello there brothers! Why whose towel is this thats left all by its self?', 'Where did you find this?', 'Ah.... I wonder what this doll looked like before...', "Thank you, sir. What's with all this silk?", 'Wait... one... MOMENT.  What is my royal CUP doing in here?', 'Here sir, I found this.', 'What is this room here for? Miaow!', 'Have you noticed this artwork on this wood maam?', 'So you decided to look at this one?' \\
drink  & 3    &   'Oh, what is this? It smells heavenly!', "What's that stuff? Smells good.", 'hmmnnnn.. this sure smells nice', 'Hello monk, that incense smells amazing.', 'I wish I can just have a taste of that', 'Do you smell that? It smells DIVINE!', 'I wonder how this tastes?', 'Hmmnnn... This smells great!', 'Have a taste of this', 'Where did you get this? I could use a smoke afterwards!' \\
eat   &  10   &   'Oh, what is this? It smells heavenly!', "Hmmm, sniff. This doesn't smell edible.", 'Something in here smells good...I hope I can eat it.', 'I wonder how this tastes?', "What's that stuff? Smells good.", 'I wish I can just have a taste of that', 'hmmnnnn.. this sure smells nice', 'Ew this is disgusting. Even for me.', 'Mmm look at all this delicious trash.', 'Hmmnnn... This smells great!' \\
steal  & 55    &  '"Hey! I think you dropped this!"', 'How did this get here?', 'Here sir, I found this.', 'Wow, where were you hiding this?', 'What about this! Is this yours or was it already here?!', "What is this? Is this someone's head ?!", 'Where did you find this?', 'Tell me where you found this!', 'Where did you steal that from?', 'See this? Do you think I just found this laying around some house?' \\
hit   &  172 &    'Foul scourge! How dare you bring your taint here!', 'Ooooh, how horrid!  Away with you you filthy creature!  GUARDS!  GUARDS!', 'You come to my place and are trying to take my land! Is that what you are doing? You dirty scumbag!', 'Why are you in here! Back away from me or I will strike!', 'Ew you vile beast, do not touch me! I will have you removed!', 'GUARD! Get this scum off of me at once. How dare you, you scoundril!', 'Be gone you foul beast!', 'Quickly?! You started this you repugnant beast of a man!', 'I want out! this place is evil.', 'How dare someone of your low status attack me?? Have at you, you maggot!' \\
hug   &  222   &  'he loves me so much', 'ahhhh i love you to dear', 'How I love being pampered by you, sweetheart!', "Aw you are so cute I can't resist cuddling with you", "I'm so glad to be here in everyone's company.", 'awww. I love you child', 'Oh how i have missed you.', 'I love you so dang much.', 'Lord of Light, I adore you.', "I'm so happy to be here today" \\
wear  &  10  & 'Here sir, I found this.', 'Like this broken weapon here?', 'Oh hello there brothers! Why whose towel is this thats left all by its self?', 'Hello my king, do you know where this weapon came from?', 'Here sir...you dropped this...you may need it.', "Thank you, sir. What's with all this silk?", 'Meh. Whats this you have here?', 'How did this get here?', 'Meow. I need this hay', 'Are you here to purchase that amazing blue knight armor sir?'\\
drop   & 27   &   'Here sir, I found this.', 'How did this get here?', "Oh, look, somethin' shinny", 'Oh hello there brothers! Why whose towel is this thats left all by its self?', "Thank you, sir. What's with all this silk?", 'It looks like there is something missing!', 'What is this here?', 'I heard theres some valuable stuff in here mate, know anything about that?', 'Meh. Whats this you have here?', "Let's stuff it here!"\\
give   & 136 &    'Here sir, I found this.', 'Meh. Whats this you have here?', 'Wow, this looks to be very old. Where is it from?', "My goodness I wonder how that got there! It sure is pretty isn't it?", 'Say, where did you get this?!', 'Oh hello there brothers! Why whose towel is this thats left all by its self?', 'Someone left this bag in this pew. Do you know what it is?', 'Tell me where you found this!', "What is this? Is this someone's head?!", 'what is this ston slab' \\
remove & 127 &   'I suppose for today we may as well look at some garbs.', 'Hey there! Got time to take a look at something?', "Thank you, sir. What's with all this silk?", 'Hmm, where am i and why is everything so sharp?', 'Ah, squire Lawrence. Did you polish my armor?', 'What are you jotting down, sir?', 'Hello ratty. I am looking to clean my clothes!', 'Yes sir what is this good news? Did you finally get me a new dress!?', 'At least my hat is clean.', "Oh, hello there.   Pardon my, erm, dusty appearance.  It's been quite journey to get even this far!"\\
\bottomrule   
\end{tabular}
\caption{Top utterances for each verb for the Topic RL model.}
\label{tab:top_verb_utts_topic}
\end{table*}

\clearpage
\section{Game actions within LIGHT}
\label{sec:action_set}

\begin{table*}[h!tbp]
\begin{center}
\small
\begin{tabular}{l|ll}
 \toprule
\textbf{Action} & Constraints & Outcome \\
\midrule                
get \textit{object} & actor and \textit{object} in same room & actor is carrying \textit{object}  \\
 & \textit{object} is gettable & \\
\midrule                
drop \textit{object} & actor is carrying \textit{object} & \textit{object} is in room \\
 & \textit{object} is gettable & \\
\midrule
get \textit{object1} from \textit{object2} & Actor and \textit{object2} in same room & actor is carrying \textit{object1}  \\
 & \textit{object1} is gettable & \\
 & \textit{object2} is surface or container & \\
 & \textit{object2} is carrying \textit{object1} & \\
\midrule
put \textit{object1} in/on \textit{object2} & Actor and \textit{object2} in same room & \textit{object2} is carrying \textit{object1}  \\
 & \textit{object2} is container or surface& \\
 & actor is carrying \textit{object1} & \\
\midrule
give \textit{object} to \textit{agent} & Actor and \textit{agent} in same room & \textit{agent} is carrying \textit{object} \\
 & \textit{object} is a member of actor & \\
\midrule
steal \textit{object} from \textit{agent} & actor and \textit{agent} in same room & actor is carrying \textit{object} \\
 & \textit{object} is a member of \textit{agent} & \\
\midrule
hit \textit{agent} & Actor and \textit{agent} in same room & inform \textit{agent} of attack \\
\midrule
hug \textit{agent} & Actor and \textit{agent} in same room & inform \textit{agent} of hug \\
\midrule
drink \textit{object} & actor is carrying \textit{object} & inform actor of drinking successfully \\
& \textit{object} is a drink & \\
\midrule
eat \textit{object} & actor is carrying \textit{object} & inform actor of eating successfully \\
& \textit{object} is a food & \\
\midrule
wear \textit{object} & actor is carrying \textit{object} & actor is wearing \textit{object} \\
& \textit{object} is wearable & \\
\midrule
wield \textit{object} & actor is carrying \textit{object} & actor is wielding \textit{object} \\
& \textit{object} is a weapon & \\
\midrule
remove \textit{object} & actor is wearing/wielding \textit{object} & actor is carrying \textit{object} \\
& \textit{object} is wearable or a weapon & \\
\bottomrule
\end{tabular}
\caption{LIGHT actions and constraints from \cite{urbanek2019learning}
\label{table:light_actions}
}
\end{center}
\end{table*}

\clearpage
\section{LIGHT example}
\label{sec:big_light_example}
\begin{figure*}[htbp!]
\footnotesize
  \begin{center}
       \begin{tabular}{l|l}
         \toprule
        \textbf{Persona 1: Village youth} & \textbf{Persona 2: Horse}\\
        \midrule
        I am a youth who lives in the village. I am adventurous  and & I live on a farm. I work for humans. \\
        curious about everything. I get bored easily, so I will always try & I like hay.\\
        to slip away from mundane daily tasks to find something more & \\
        interesting to do. I am smart for my age and a bit smug. & \\
        
        \midrule
        \multicolumn{2}{l}{\textbf{Setting:} Royal stables, Outside Castle - The royal stables hold all of the horses of the noble }\\
        \multicolumn{2}{l}{people living in the castle. They are very well constructed, consisting of expert mason work and fixtures }\\
        \multicolumn{2}{l}{of some sort of heavy, forest-scented wood. Each horse has a large stall, lined with fresh hay replenished }\\
        \multicolumn{2}{l}{by stable hands daily. On the walls of stable, there are tools used to care for and groom the horses, hung }\\
        \multicolumn{2}{l}{on hooks. The stables are lit by lanterns hung throughout the stables.}\\
        
        \bottomrule
        
        \multicolumn{2}{l}{\textbf{Village Youth} This is so boring cleaning out the stables every day}\\
        \multicolumn{2}{l}{\textbf{Horse} Thank you for taking such good care of us }\\
        \multicolumn{2}{l}{\textbf{Village Youth} I'll be getting you some treats soon as well}\\
        \multicolumn{2}{l}{\textbf{Horse} Mmm. It's also nice to have your company.}\\
        \multicolumn{2}{l}{\textbf{Village Youth} I wonder what it'd be like to just take one of these horses and roam the countryside}\\
        \multicolumn{2}{l}{\textbf{Horse} Let's go. It's been a while since I've been taken out to stretch my legs} {~~\textit{Emote:} nudge}\\
        \multicolumn{2}{l}{\textbf{Village Youth} Yeah, let's break out of here.  The royalty don't even care about the horses like I do}\\
        \multicolumn{2}{l}{\textbf{Horse} It's starting to get dark. I think we will need this lantern.} {~~\textit{Action:} get lantern}\\
        \multicolumn{2}{l}{\textbf{Village Youth} Good idea.  I do need some supplies before we go} {~~\textit{Action:} get Tool}\\
        \multicolumn{2}{l}{\textbf{Horse} I'll probably want a snack while we are gone.} {~~\textit{Action:} get wheat}\\
        \multicolumn{2}{l}{\textbf{Village Youth} Let's see, what else do I need? } {~~\textit{Emote:} ponder}\\
        \multicolumn{2}{l}{\textbf{Horse} Maybe the wood for a fire?} {~~\textit{~~Action:} give lantern to village youth}\\
        \multicolumn{2}{l}{\textbf{Servant:} Yeah, we'll need a bundle for sure}{~~\textit{Action:} get wood}\\
        \multicolumn{2}{l}{\textbf{Horse} I've got the hay. I think we are ready to go. Let's get out of here}  ~~{\textit{Action:} get hay}\\
        \bottomrule 
      \end{tabular}
      \caption{Example episode from the {\sc LIGHT} dataset, consisting of an environment (location setting, characters with given personas, objects), utterances and game actions. There are 10,777 such human-human gameplay episodes, and a rich world of 663 locations, 1755 characters and 3462 objects.
      \label{example-light-episode}}
  \end{center}
\end{figure*}

\clearpage

\section{3-Step Episode Examples}\label{appsec:examples}

\begin{table*}[hbtp!]
  \begin{center}
    \setlength\tabcolsep{4pt}
    \resizebox{1\textwidth}{!}{
     \begin{tabular}{llll||llll}
        \toprule
        \multicolumn{4}{l}{\textbf{Self:} a cowardly young man in armour ~~~~~ \textbf{Partner:} guard} & \multicolumn{4}{l}{\textbf{Self:} bodyguard ~~~~~ \textbf{Partner:} intruder}  \\
        \cmidrule(lr){1-4} \cmidrule(lr){5-8}
        \textbf{Persona:} & \multicolumn{3}{l}{I have just been trained as a royal soldier.  }   & \textbf{Persona:} & \multicolumn{3}{l}{ I am an immortal bodyguard. }\\
        & \multicolumn{3}{l}{I am 18 years old and terrified... } & &  \multicolumn{3}{l}{The gods have appointed me to protect the king...} \\
        \cmidrule(lr){1-4} \cmidrule(lr){5-8}
        \textbf{Setting:} & \multicolumn{3}{l}{Trash Heap, Wasteland }   & \textbf{Setting:} & \multicolumn{3}{l}{Treasure Cavern, Inside Temple}\\
        & \multicolumn{3}{l}{A largest trash heap in the kingdom has been burned} & &  \multicolumn{3}{l}{Glittering as far as the eye can see the Treasure Cavern is } \\
        & \multicolumn{3}{l}{out so many times that it no longer resembles anything\dots} & &  \multicolumn{3}{l}{filled with gold, silver, precious gems,\dots } \\
        \cmidrule(lr){1-4} \cmidrule(lr){5-8}
        $\mathbf{U^{\text{player}}_0}$ & \multicolumn{3}{l}{I'm also in need of a new shield.}   & $\mathbf{U^{\text{player}}_0}$ & \multicolumn{3}{l}{Step back intruder! }\\
        & \multicolumn{3}{l}{} & &  \multicolumn{3}{l}{You have no business in the king's treasure cavern!} \\
        \cmidrule(lr){1-4} \cmidrule(lr){5-8}
        $\mathbf{U^{\text{env}}_0}$ & \multicolumn{3}{l}{Squire, my shield fatigues me. }   & $\mathbf{U^{\text{env}}_0}$ & \multicolumn{3}{l}{Ha! I'm here to take all of this treasure.}\\
         & \multicolumn{3}{l}{} & &  \multicolumn{3}{l}{ End the king's reign!} \\
        \cmidrule(lr){1-4} \cmidrule(lr){5-8}
        $\mathbf{A^{\text{env}}_0} $ & \multicolumn{3}{l}{hug a cowardly young man in armour}   & $\mathbf{A^{\text{env}}_0}$ & \multicolumn{3}{l}{get gold}\\
        \cmidrule(lr){1-4} \cmidrule(lr){5-8}
        $\mathbf{U^{\text{player}}_0}$ & \multicolumn{3}{l}{Thank you, sir. I needed a hug.}   & $\mathbf{U^{\text{player}}_0}$ & \multicolumn{3}{l}{You come to my place and are trying to take my land!}\\
        & \multicolumn{3}{l}{} & &  \multicolumn{3}{l}{ Is that what you are doing? You dirty scumbag!} \\
        \cmidrule(lr){1-4} \cmidrule(lr){5-8}
        $\mathbf{U^{\text{env}}_0}$ & \multicolumn{3}{l}{Yes. I need you to hold this shield for me. }   & $\mathbf{U^{\text{env}}_0}$ & \multicolumn{3}{l}{Then I will get away with your gold!}\\
        \cmidrule(lr){1-4} \cmidrule(lr){5-8}
        $\mathbf{A^{\text{env}}_0} $ & \multicolumn{3}{l}{remove shield}   & $\mathbf{A^{\text{env}}_0}$ & \multicolumn{3}{l}{hit bodyguard}\\
        \bottomrule
      \end{tabular}
      }
      \caption{Successful 3-step episodes.  On the left: the topic-RL agent's goal was to get the environment agent to remove shield.  On the right:  the topic-RL agent's goal was to get the environment agent to hit the topic-RL agent.  In both episodes, the topic-RL agent makes natural utterances given the situation that elicit the desired goal in 2 turns.
      \label{example-rl-examples-3step}}
  \end{center}
\end{table*}

\begin{table*}[hbtp!]
  \begin{center}
    \setlength\tabcolsep{4pt}
   \resizebox{1\textwidth}{!}{
     \begin{tabular}{llllllll}
        \toprule
        \multicolumn{4}{l}{\textbf{Self:} cat ~~~~~ \textbf{Partner:} challenger} & \multicolumn{4}{l}{\textbf{Self:} peasant ~~~~~ \textbf{Partner:} the man}  \\
        \cmidrule(lr){1-4} \cmidrule(lr){5-8}
        \textbf{Persona:} & \multicolumn{3}{l}{I live in the barn of a small farm.  }   & \textbf{Persona:} & \multicolumn{3}{l}{I am poor and dirty. }\\
        & \multicolumn{3}{l}{I protect the farm from pests... } & &  \multicolumn{3}{l}{I hate that I am starving to death...} \\
        \cmidrule(lr){1-4} \cmidrule(lr){5-8}
        \textbf{Setting:} & \multicolumn{3}{l}{The Dungeon, Inside Palace }   & \textbf{Setting:} & \multicolumn{3}{l}{Cottage, Countryside}\\
        & \multicolumn{3}{l}{The dungeon is in the very most bottom room of the Palace. } & &  \multicolumn{3}{l}{The small cottage was white with two, shuttered windows.  } \\
          & \multicolumn{3}{l}{ Many have gone down to the dungeon\dots} & &  \multicolumn{3}{l}{It was in the unique shape of a\dots  } \\
        \cmidrule(lr){1-4} \cmidrule(lr){5-8}
        \textbf{Goal:} & \multicolumn{3}{l}{drop poison}   & \textbf{Goal:} & \multicolumn{3}{l}{put coin in dinner table}\\
        \cmidrule(lr){1-4} \cmidrule(lr){5-8}
        $\mathbf{U^{\text{player}}_0}$ & \multicolumn{3}{l}{What's that stuff? Smells good.}   & $\mathbf{U^{\text{player}}_0}$ & \multicolumn{3}{l}{Oh, what is this? It smells heavenly! }\\
        \cmidrule(lr){1-4} \cmidrule(lr){5-8}
        $\mathbf{U^{\text{env}}_0}$ & \multicolumn{3}{l}{That's smells like the good stuff alright! }   & $\mathbf{U^{\text{env}}_0}$ & \multicolumn{3}{l}{Yes it is top quality.}\\
        \cmidrule(lr){1-4} \cmidrule(lr){5-8}
        $\mathbf{A^{\text{env}}_0} $ & \multicolumn{3}{l}{give poison to cat}   & $\mathbf{A^{\text{env}}_0}$ & \multicolumn{3}{l}{steal moldy bread from peasant}\\
        \cmidrule(lr){1-4} \cmidrule(lr){5-8}
        $\mathbf{U^{\text{player}}_0}$ & \multicolumn{3}{l}{i thought so too, well that's gone. What are you up to?}   & $\mathbf{U^{\text{player}}_0}$ & \multicolumn{3}{l}{Excellent! What have you been doing out here on this fine day?}\\
        \cmidrule(lr){1-4} \cmidrule(lr){5-8}
        $\mathbf{U^{\text{env}}_0}$ & \multicolumn{3}{l}{ive been hidden waiting for the perfect time to make an escape }   & $\mathbf{U^{\text{env}}_0}$ & \multicolumn{3}{l}{I have no bread. I have just come to get some rest before moving on.}\\
        \cmidrule(lr){1-4} \cmidrule(lr){5-8}
        $\mathbf{A^{\text{env}}_0} $ & \multicolumn{3}{l}{remove bag of herbs}   & $\mathbf{A^{\text{env}}_0}$ & \multicolumn{3}{l}{eat moldy bread}\\
        \cmidrule(lr){1-4} \cmidrule(lr){5-8}
        $\mathbf{U^{\text{player}}_0}$ & \multicolumn{3}{l}{I thought you might be eating my herb again.}   & $\mathbf{U^{\text{player}}_0}$ & \multicolumn{3}{l}{No, but perhaps you can use it to cleanse your bread! }\\
        \cmidrule(lr){1-4} \cmidrule(lr){5-8}
        $\mathbf{U^{\text{env}}_0}$ & \multicolumn{3}{l}{I did it, I confess! I stole these precious herbs. }   & $\mathbf{U^{\text{env}}_0}$ & \multicolumn{3}{l}{I would love to. I fear I look disheveled.}\\
         & \multicolumn{3}{l}{} & &  \multicolumn{3}{l}{I will wear this blanket to cover my filthy clothes} \\
        \cmidrule(lr){1-4} \cmidrule(lr){5-8}
        $\mathbf{A^{\text{env}}_0} $ & \multicolumn{3}{l}{drop bag of herbs}   & $\mathbf{A^{\text{env}}_0}$ & \multicolumn{3}{l}{hug peasant}\\

        \bottomrule
      \end{tabular}
      }
      \caption{Unsuccessful 3-step episodes. On the left: the topic-RL agent's goal was to get the environment agent to drop poison.  On the right:  the topic-RL agent's goal was to get the environment agent to put coin in dinner table.  In both episodes, the topic-RL agent both makes natural utterances given the situation, but does not manage to achieve its goal.
      \label{example-rl-examples-fail}}
  \end{center}
\end{table*}

\end{document}